\documentclass[11pt]{article}
\usepackage{eamt20}
\usepackage{times}
\usepackage{url}
\usepackage{latexsym}
\usepackage[small,bf]{caption} 
\usepackage{times}
\usepackage{latexsym}

\usepackage{microtype}
\usepackage[T1]{fontenc}
\usepackage{tipa}
\usepackage{times}
\usepackage{graphicx} 
\usepackage{latexsym}
\usepackage{amsthm}
\usepackage{amssymb}
\usepackage[titletoc,title]{appendix}
\usepackage{soul}
\usepackage{tabularx}
\usepackage{graphicx}
\usepackage{caption}
\usepackage{amsthm}
\theoremstyle{definition}

\theoremstyle{remark}

\usepackage{url}
\usepackage{makecell}
\RequirePackage{filecontents}
\usepackage[strict]{changepage}
\usepackage{array,multirow}
\usepackage{booktabs}
\usepackage{CJKutf8}
\usepackage{makecell} 
\usepackage{url}
\usepackage{booktabs}
\usepackage{enumitem}
\setlist{nosep}
\usepackage[compact]{titlesec}

\title{Using Interlinear Glosses as Pivot in Low-Resource
  \\ Multilingual Machine Translation} 

\author{
  Zhong Zhou\\
  {\small Carnegie Mellon University}\\
  {\small \tt zhongzhou@cmu.edu}
  \\\And
  Lori Levin\\
  {\small Carnegie Mellon University}\\
  {\small \tt lsl@cs.cmu.edu}
  \\\And
  David R. Mortensen\\
  {\small Carnegie Mellon University}\\
  {\small \tt dmortens@cs.cmu.edu}
  \\\And
  Alex Waibel\\
  \small Carnegie Mellon University\\
  \small Karlsruhe Institute of Technology\\
  {\small \tt alex@waibel.com }
}

\date{}

\begin{document}
\maketitle

\begin{abstract}
  We demonstrate a new approach to Neural Machine Translation (NMT) for low-resource languages using a ubiquitous linguistic resource, \textit{Interlinear Glossed Text} (IGT). IGT represents a non-English sentence as a sequence of English lemmas and morpheme labels. As such, it can serve as a pivot or interlingua for NMT. Our contribution is four-fold. Firstly, we pool IGT for 1,497 languages in ODIN (54,545 glosses) and 70,918 glosses in Arapaho and train a gloss-to-target NMT system from IGT to English, with a BLEU score of 25.94. We introduce a multilingual NMT model that tags all glossed text with gloss-source language tags and train a universal system with shared attention across 1,497 languages. Secondly, we use the IGT gloss-to-target translation as a key step in an English-Turkish MT system trained on only 865 lines from ODIN. Thirdly, we we present five metrics for evaluating extremely low-resource translation when BLEU is no longer sufficient and evaluate the Turkish low-resource system using BLEU and also using accuracy of matching nouns, verbs, agreement, tense, and spurious repetition, showing large improvements. 
\end{abstract}

\section{Introduction}
\label{sec:introduction}
\begin{figure}
  \centering
  \includegraphics[width=2.7in]{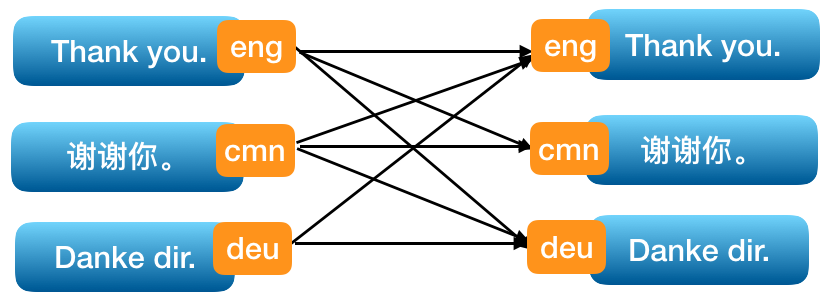}
  \caption{Multilingual NMT.}
  \label{figure:multi}
\end{figure}
\begin{figure}
  \centering
  \includegraphics[width=2.7in]{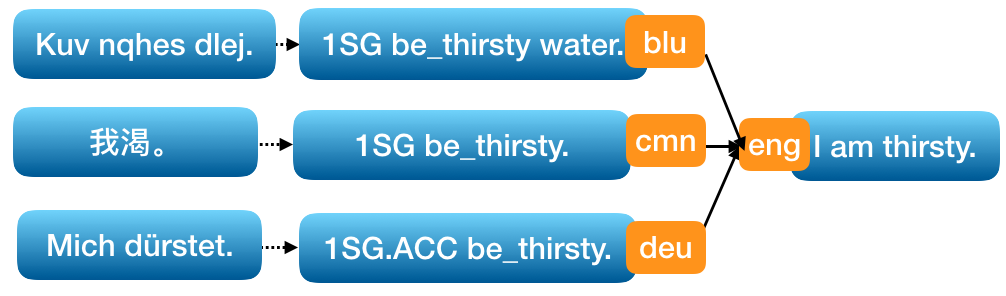}
  \caption{Using interlinear glosses as pivot to translate in multilingual NMT in Hmong, Chinese, and German.}
  \label{figure:inl}
\end{figure}
\begin{table*}[t]
\hspace*{-0.5cm} 
  \centering
  \begin{tabular}{ p{5.1 cm}  p{10.1 cm} } \hline
   \toprule
    Data & Example \\ 
    \midrule
    Source language (German) & Ich sah ihm den Film gefallen.\\
    Interlinear Gloss w/ target-lemma & I saw he.DAT the film.ACC like. \\
    Target language (English) & I saw him like the film. \\
    \midrule
    Source language (Hmong) & Nwg yeej qhuas nwg. \\ 
    Interlinear gloss w/ target-lemma & 3SG always praise 3SG. \\ 
    Target language (English) & He always praises himself. \\ 
    \midrule
    Source language (Arapaho) & Niine'etii3i' teesiihi' coo'oteyou'uHohootino' nenee3i' neeyeicii. \\ 
    Interlinear gloss w/ target-lemma & live(at)-3PL  on/over-ADV IC.hill(y)-0.PLtree-NA.PL IC.it is-3PL timber. \\ 
    Target language (English) & Trees make the woods. \\
    \bottomrule
  \end{tabular}
  \caption{Examples of interlinear glosses in different source languages. 
  }
  \label{table:examplesInterlinearGloss}
\end{table*}

Machine polyglotism, training a universal NMT system with a shared attention through implicit parameter sharing, is very helpful low-resource settings \cite{firat2016multi,zoph2016multi,dong2015multi,gillick2016multilingual,al2013polyglot,zhong2018massively,tsvetkov2016polyglot}. However, there is still a large disparity between translation quality in high-resource and low-resource settings, even when the model is well-tuned \cite{koehn2017six,sennrich2019revisiting,nordhoff2013glottolog}. Indeed, there is room for creativity 
in low-resource scenarios.   

Morphological analysis is useful in reducing word sparsity in low-resource languages \cite{habash2006arabic,lee2004morphological,hajic2000machine}.   Toward that end, we leverage a linguistic resource, \textit{Interlinear Glossed Text} (IGT)~\cite{Leipzig} as shown in Table~\ref{table:translationSequence} \cite{samardzic2015automatic,moeller2018automatic}. We propose to use interlinear gloss as a pivot to address the harder problem of morphological complexity and source of data sparsity in multilingual NMT. We combine the benefits of both multilingual NMT and linguistic information through the use of interlinear glosses as a pivot representation. Our contribution is four-fold. 
\begin{enumerate}
  \item We present our multilingual model using a single attention in translating from interlinear glosses into a target language. Our best multilingual NMT result achieves a BLEU score of 25.94, +5.65 above a single-source single-target NMT baseline.
  \item We present two linguistic datasets that we normalized, cleaned and filtered: the cleaned ODIN dataset includes 54,545 lines of IGT in 1,496 languages \cite{lewis2010developing,xia2014enriching}, and the cleaned Arapaho dataset includes 70,918 lines of IGT \cite{cowell2012speech,wagner2016applying}. 
  \item We present a three-step approach for extremely low-resource translation.  We demonstrate it by training only on 865 lines of data.  
  \begin{enumerate}
    \item We use a morphological analyzer to automatically generate interlinear glosses with source lemma. 
    \item We translate the source lemma into target lemma in through alignments trained from parallel data.  
    \item We translate from interlinear glosses with target lemma to target language by using the gloss-to-target multilingual NMT developed in 1 presented above. 
\end{enumerate}
  \item We present five metrics for evaluating extremely low-resource translation when BLEU no longer suffices. Our system using interlinear glosses achieves an improvement of +44.44\% in Noun-Verb Agreements, raising fluency.
\end{enumerate}
We present our cleaned data followed by gloss-to-target models and our three-step Turkish-English NMT in Section~\ref{sec:data} and~\ref{sec:methodology}. 
We evaluate in Section~\ref{sec:results}. 
\begin{table*}[t]
  \centering
  \begin{tabular}{p{6.3 cm} p{7.0 cm} } 
    \toprule
    Data & Example \\ 
    \midrule
    \textit{1}. Source language (Turkish) & Kadin dans ediyor. \\ 
    \textit{2}. Interlinear gloss with source-lemma & Kadin.NOM dance ediyor-AOR.3.SG. \\
    \textit{3}. Interlinear gloss with target-lemma & Woman.NOM dance do-AOR.3.SG. \\
    \textit{4}. Target language (English) & The woman dances. \\ 
    \midrule
    \textit{1}. Source language (Turkish) & Adam kadin-i gör-dü. \\
    \textit{2}. Interlinear gloss with source-lemma & Adam.NOM kadin-ACC gör-AOR.3.SG. \\
    \textit{3}. Interlinear gloss with target-lemma & Man.NOM woman-ACC see-PST.3.SG. \\
    \textit{4}. Target language (English) & The man saw the woman. \\
    \bottomrule
  \end{tabular}
  \caption{Examples of the translation sequence using interlinear glosses.
  }
  \label{table:translationSequence}
\end{table*}
\begin{table}[t]
  \centering
  \begin{tabular}{ p{1.2 cm} p{6 cm} } 
    \toprule
    Notation & Meaning in translation sequence \\ \midrule
    \textit{1} & Source language (Turkish) text \\
    \textit{2} & Interlinear gloss with source-lemma \\
    \textit{3} & Interlinear gloss with target-lemma \\
    \textit{4} & Target language (English) text \\
    \bottomrule
  \end{tabular}
  \caption{Notation used in the translation sequence.
  }
  \label{table:notation}
\end{table}
\section{Related Works}
\label{sec:relatedWorks}
\subsection{Multilingual Neural Machine Translation}
Multilingual NMT's objective is to translate
from any of $N$ input languages to any of 
$M$ output languages \cite {firat2016multi,zoph2016multi,dong2015multi,gillick2016multilingual,al2013polyglot,tsvetkov2016polyglot}. Many multilingual NMT systems work on 
a universal model with a shared attention mechanism
with Byte-Pair Encoding (BPE) \cite{johnson2017google,ha2016toward,zhong2018massively}. Its simplicity and implicit 
parameter sharing helps with low-resource translation and zero-shot translation \cite{johnson2017google,firat2016multi}.

\subsection{Morpheme-Level Machine Translation}
To build robustness 
\cite{chaudhary2018adapting,cotterell2015morphological,wu2016google}, researchers work on
character-level, byte-level \cite{gillick2016multilingual,ling2015character,chung2016character,tiedemann2012character}, and BPE-level \cite{sennrich2016neural,burlot2017word} translation. 
\begin{table*}[th!]
  \centering
  \begin{tabular}{ p{3cm } p{4.5 cm} p{6.7 cm}} 
    \toprule
    Normalized Gloss & Meaning of the Abbreviations & Glosses in the Turkish Odin Data \\ 
    \midrule
    NMLZ & Nominalizer & NML, NOMZ, FNom, NOML \\
    PRS & Present tense & PRES, PR, pres, Pres, PRESENT \\
    PST & Past tense & PA, Pst, PST, Past, pst, PAST, PT, PTS, REPPAST, PST1S, past \\
    ABL & Ablative & Abl, Abli, abl, ABL \\
    ADV & Adverb(ial) & ADVL, Adv \\
    RPRT & Reported Past tense & ReportedPast, REPPAST \\
    \bottomrule
  \end{tabular}
  \caption{Examples of the normalization mapping created for the Turkish ODIN data.
  }
  \label{table:examplesDictNorm1}
\end{table*}
\begin{table*}[th!]
  \centering
  \begin{tabular}{p{2.5cm } p{5 cm} p{5.7 cm}} 
    \toprule
    Tags & Meaning of the Abbreviations & Sets of Normalized Glosses Included \\ 
    \midrule
    P1pl & 1st person plural possessive & 1, PL, POSS \\
    A1sg & 1st person singular & 1, SG \\
    Reflex & Reflexive Pronoun & REFL \\ 
    NarrPart & Evidential participle & EVID, PTCP \\\
    AorPart & Aorist participle & AOR, PTCP \\
    PresPart & Present participle & PRS, PTCP \\
    \bottomrule
  \end{tabular}
  \caption{Examples of the normalization mapping created for the outputs from the Turkish morphological analyzer. 
  }
  \label{table:examplesDictNorm2}
\end{table*}
\begin{table*}[t]
  \centering
  \begin{tabular}{ p{5.5 cm} p{9.0 cm}} 
    \toprule
    Interlinear Gloss w/ Target-lemma & Example \\ 
    \midrule
    Before normalization & Ahmet self-3.sg-ACC very admire-Progr.-Rep.Past. \\
    After normalization & Ahmet self-3.SG-ACC very admire-PROG-Rep.PST. \\ \midrule
    Before normalization & Woman.NOM dance do-AOR.3SG. \\
    After normalization  & Woman.NOM dance do-AOR.SG.3. \\ 
    \midrule
    Before normalization & Man.NOM woman-ACC see-PAST.3SG. \\
    After normalization & Man.NOM woman-ACC see-PST.SG.3. \\ 
    \bottomrule
  \end{tabular}
  \caption{Examples of the normalization of glosses from the Turkish ODIN data.
  }
  \label{table:normalization}
\end{table*}
Morpheme-level translation
allows words to share embedding while allowing variation
in meanings \cite{cotterell2015morphological,chaudhary2018adapting,renduchintala2019character,passban2018improving,dalvi2017understanding},
shrinks the vocabulary size 
introduces smoothing \cite{goldwater2005improving},  
and makes fine-grained
correction \cite{stroppa2006example,matthews2018using}. 
\begin{table*}[t]
  \centering
  \begin{tabular}{ p{3.6 cm} p{11.0 cm}} 
    \toprule
    Analyzer Outputs & Example \\ 
    \midrule
    Before normalization & Kadi+A3sg+Pnon+Nom dans+A3sg+Pnon+Nom et+Prog1+A3sg. \\ 
    After normalization & Kadin.3.SG.NPOSS.NOM dans.3.SG.NPOSS.NOM ediyor-PROG.3.SG. \\ 
    \midrule
    Before normalization & Adam+A3pl+Pnon+Nom kadi+A3sg+Pnon+Acc gör+Past+A3sg. \\ 
    After normalization  & Adam.3.SG.NPOSS.NOM kadi.3.SG.NPOSS.ACC gör-PST.3.SG. \\
    \midrule
    Before normalization & Ali+A3sg+Pnon+Nom hakkinda+A3sg+P3sg+Loc Ahmet+Prop+A3sg+Pnon+Nom ne düünüyor+A3sg+Pnon+Nom?  \\
    After normalization & Ali.3.SG.NPOSS.NOM hakkinda.3.SG.POSS.LOC Ahmet.3.SG.NPOSS.NOM ne düünüyor.3.SG.NPOSS.NOM? \\ 
    \bottomrule
  \end{tabular}
  \caption{Examples of the normalization process for the output of the Turkish morphological analyzer. 
  }
  \label{table:normalizationMorphologicalAnalyzer}
\end{table*}

\subsection{Factored Machine Translation} 
Factored models translate a
composition of annotations including word,
lemma, part-of-speech, morphology, and word class into the
target language \cite{koehn2007factored,yeniterzi2010syntax}. In the era of NMT, morphological information and grammatical decomposition that are produced by a morphological analyzer are employed \cite{garcia2016factored,burlot2017word,hokamp2017ensembling}.  

\subsection{Interlinear Gloss Generation} 
Interlinear gloss is a linguistic representation of morphosyntactic categories and cross-linguistic lexical relations \cite{samardzic2015automatic,moeller2018automatic}.  IGT is used in linguistic publications and field notes to communicate technical facts about languages that the reader might not speak, or to convey a particular linguistic analysis to the reader. Typically, there are three lines to an IGT. The first line consists of text segments in an \textit{object language}, which we call the \textit{source language} in this paper.  The third line is a fluent translation in the \textit{metalanguage}, which we call the \textit{target language}. In our work, the target language (metalanguage) is always English.  The source (object) languages are the 1,496 languages of the ODIN \cite{lewis2010developing,xia2014enriching} database plus Arapaho, for which a large collection of field notes has been published~\cite{cowell2012speech}.  In the middle (interlinear) line of an IGT, each object language word is represented as an English (metalanguage) lemma and labels for the non-English morphemes.  For our work, we add a fourth line:  a source-language lemma with morpheme labels.   To illustrate with an example from Arapaho, an endangered language that is spoken in the United States by less than 200 people.~\cite{cowell2012speech}, ``hohoot nii3eihit" means ``a tree is nice". The interlinear gloss with source-language lemmas is ``hohoot nii3eihit-3.S" and the interlinear gloss with target-language lemmas is ``Tree good-3.S". 

A benefit of IGT is that there is a
one-to-one mapping between each segment of the source sentence to the gloss \cite{samardzic2015automatic}. Researchers have tried to generate interlinear glosses automatically by using supervised POS tagging, word disambiguation and a dictionary \cite{samardzic2015automatic}, and using conditional random fields and active learning \cite{moeller2018automatic}. 

\section{Data} 
\label{sec:data}
\begin{table*}[t]
  \centering
  \begin{tabular}{p{4.1 cm} p{11 cm}} 
    \toprule
    Analyzer Outputs & Example \\ 
    \midrule
    Before normalization & Kadi+A3sg+Pnon+Nom dans+A3sg+Pnon+Nom et+Prog1+A3sg. \\ 
    After normalization & Kadin.3.SG.NPOSS.NOM dans.3.SG.NPOSS.NOM ediyor-PROG.3.SG. \\ 
    After using a dictionary & Woman.3.SG.NPOSS.NOM dance.3.SG.NPOSS.NOM be-PROG.3.SG. \\ 
    Reference in ODIN & Woman.NOM dance do-AOR.3.SG. \\ 
    \midrule
    Before normalization & Adam+A3pl+Pnon+Nom kadi+A3sg+Pnon+Acc gör+Past+A3sg. \\ 
    After normalization & Adam.3.SG.NPOSS.NOM kadi.3.SG.NPOSS.ACC gör-PST.3.SG. \\ 
    After using a dictionary & Man.3.SG.NPOSS.NOM woman.3.SG.NPOSS.ACC see-PST.3.SG. \\ 
    Reference in ODIN & Man.NOM woman-ACC see-PST.3.SG. \\ 
    \bottomrule
  \end{tabular}
  \caption{Examples of interlinear gloss generation (\textit{1}$\rightarrow$\textit{2}$\rightarrow$\textit{3}) from the output from the Turkish morphological analyzer. Notation of the translation sequence follows from Table~\ref{table:notation}.
  }
  \label{table:normalization2}
\end{table*}
\begin{table*}[t]
  \centering
  \begin{tabular}{p{6.1 cm} p{4.7 cm} p{4.7 cm} } 
    \toprule
    Interlinear Gloss w/ Target-lemma & NMT Result in Target Language & Reference Target Sentence\\ 
    \midrule
    Peter and Mary that/those not came-3SG/3PL	
    & Peter and Mary , he didn't come.	
    & Peter and Mary, they didn't come.
    \\ 
    PERF.AV-buy NOM-man ERG-fish DAT-store	
    & The man bought fish at the store. 
    & The man bought fish at thestore'.
    \\ 
    AGR-do-make-ASP that waterpot AGR-fall-ASP	
    & The girl made that waterpot fall.	
    & The girl made the waterpot fall.
    \\ 
    \bottomrule
  \end{tabular}
  \caption{Examples of Gloss-to-Target (\textit{3}$\rightarrow$\textit{4} in Table~\ref{table:notation}) NMT translation results. The source is the interlinear gloss with target(English)-lemma and the target is the fluent English. Notation of the translation sequence follows from Table~\ref{table:notation}. Note that the ODIN dataset is not clean, and the second example above is a case where two words are concatenated together without space followed by a unnecessary punctuation symbol. This example serves to show that our NMT output automatically correct typos in producing fluent target(English) sentence.}
  \label{table:3to4}
\end{table*}
\begin{table*}[t]
  \small
  \centering
  \begin{tabular}{p{1.1 cm} p{1.1 cm} p{1.1 cm} p{1.8 cm} p{1.1 cm} p{2.1 cm} p{2.7 cm}} 
   \toprule
    Model & \textit{Turkish} & \textit{ODIN} & \textit{ODIN\_multi} & \textit{Arapaho} & \textit{ODIN+Arapaho} & \textit{ODIN+Arapaho\_multi} \\ 
    Type & single & single & multilingual & single & single & multilingual\\
    \midrule
    Data & 865 & 54,545 & 54,545 & 70,918 & 125,463 & 125,463 \\
    BLEU & 0.0 & 20.29 
    & \textbf{23.05} 
    & 22.68 
    & \textbf{25.94} 
    & \textbf{25.85} 
    \\
    \bottomrule
  \end{tabular}
  \caption{BLEU scores in gloss-to-target translation using ODIN and the Arapaho dataset. If "Type" is "single", it is trained using a single-source single-target NMT; if it is "multilingual", it is trained using a multilingual NMT. }
  \label{table:odin}
\end{table*}
\begin{table*}[t]
  \small
  \centering
  \begin{tabular}{p{4.2cm} p{1.3 cm} p{1.3 cm} p{1 cm} p{1.2 cm} p{1 cm} p{0.9cm}}
    \toprule
    Model & \textit{Baseline1} & \textit{Baseline2} & \multicolumn{2}{l}{\textit{IGT\_src}} &  
    \multicolumn{2}{l}{\textit{IGT\_tgt}}\\ 
    Translation Sequence & 1$\rightarrow$4 & 1$\rightarrow$4* & \multicolumn{2}{l}{1$\rightarrow$2$\rightarrow$4} & \multicolumn{2}{l}{1$\rightarrow$2$\rightarrow$3$\rightarrow$4} \\
    \midrule
    Data used & 865 & 58473 & \multicolumn{2}{l}{865} & \multicolumn{2}{l}{865} \\
    Supplemental resources & - & - & \multicolumn{2}{l}{Analyzer} & \multicolumn{2}{c}{Analyzer, alignments} \\ 
    \midrule
    Noun-match accuracy & 15.96 & 5.07 & 20.50 & (+4.54) & \textbf{36.11} & (+20.15) \\ 
    Verb-match accuracy & 4.63 & 7.87 & 9.72 & (+5.09) & \textbf{20.37} & (+15.74) \\ 
    Subject-verb agreement accuracy & 45.37 & 64.81 & 37.96 & (-7.41) & \textbf{89.81} & (+44.44) \\
    Tense-match accuracy & 16.67 & 40.74 & 22.22 & (+5.55) & \textbf{43.52} & (+26.85) \\ 
    Non-repetition metric & 89.24 & 96.95 & 92.46 & (+3.22) & \textbf{99.27} & (+10.03) \\ 
    4-gram BLEU & 3.05 & 2.39 & \textbf{5.08} & (+2.03) & 4.74 & (+1.66) \\
    1-gram BLEU & 21.30 & 12.3 & \textbf{28.0} & (+6.70) & 21.50 & (+0.20) \\
    \bottomrule
  \end{tabular}
  \caption{Evaluation of different translation sequences with notations from Table~\ref{table:notation}. \textit{Baseline2} uses additional 57,608 lines of Turkish-English parallel data.  
  }
  \label{table:quantitative}
\end{table*}
\begin{table*}[t]
  \small
  \centering
  \begin{tabular}{p{1.6 cm} >{\raggedright\arraybackslash}p{1.6 cm}  >{\raggedright\arraybackslash}p{2 cm} >{\raggedright\arraybackslash}p{1.7 cm} >{\raggedright\arraybackslash}p{1.8 cm} >{\raggedright\arraybackslash}p{3 cm} >{\raggedright\arraybackslash}p{1.8 cm}} 
    \toprule
    Source Sentence & Gold & \textit{\textit{Baseline1}} & \textit{\textit{Baseline2}} &
    \textit{\textit{IGT\_src}} &
    \textit{\textit{Generation}} & \textit{\textit{IGT\_tgt}}\\ Sequence &  & 1$\rightarrow$4 & 1$\rightarrow$4* & 1$\rightarrow$2$\rightarrow$4 & 1$\rightarrow$2$\rightarrow$3 &  1$\rightarrow$2$\rightarrow$3$\rightarrow$4 \\
    \midrule
    Problemi çöz-mek zor-dur.
    & To solve the problem is difficult.
    & Ali read the book.
    & As for this book , it is known that it is known as a result. 
    & As for the book , the book .
    & Solv-3.SG.ACC the is difficult-3.SG.COP.PRS.
    & The fact that it is difficult for that.
    \\ 
    \midrule
    Fatma bu kitabkimin yazd gn sanyor.
    & Who does Fatma think wrote this book.
    & As for Fatma knows that I one.
    & As for Fatma , I have a house. 
    & As for Fatma , Fatma knows that I left .
    & Fatma-3.SG.NOM this-DET kitabkimin ATATURK-3.SG.NOM wrote-3.SG.NOM it-3.SG.NOM.
    & As for Fatma, it is possible that he wrote this.
    \\ 
    \midrule
    Adam cocuga top verdi.
    & The man gave the child a ball.
    & As for the book, the book, the book, the book.
    & Ahmet read the book.
    & the girl that the book .
    & Man-1.3.SG.NOM.POSS child-3.SG.DAT ball-3.SG.NOM.
    & The man's child is the ball.
    \\ 
    \bottomrule
  \end{tabular}
  \caption{Qualitative Evaluation. All experiments except the starred \textit{Baseline2} use 865 lines of training data. \textit{Baseline2} uses additional 57608 lines of parallel data. Notation of the translation sequence follows from Table~\ref{table:notation}.
  }
  \label{table:qualitative}
\end{table*}

\subsection{Newly Cleaned Datasets} 
We present two linguistic datasets that we have cleaned and partially normalized: the partially normalized ODIN dataset that includes 54,545 lines of IGT in 1,496 languages \cite{lewis2010developing,xia2014enriching}, and the cleaned Arapaho dataset that includes 70,918 lines of IGT \cite{cowell2012speech,wagner2016applying}. 

ODIN is a unique multilingual database of IGT that was scraped from the Web.    The IGTs in ODIN come from many different publications with different standards for morpheme labels. A consequence of the diversity is that morpheme labels in ODIN are not standardized. For example, ``singular'' may be "S", "SG", or "SING", and when combined with "3" for third person, there may or may not be a delimiter, resulting such diverse labels as "3S", "3.s", "3SG", and "3.sing".    

In order to reduce the sparsity and diversity of morpheme labels, we normalized them according to the Leizig conventions \cite{lehmann1982directions,croft2002typology} (preferred), and the Unimorph conventions \cite{unimorph,kirov2016very,sylak2015language} for labels not covered by the Leipzig conventions.  For the work presented here, we normalized only those morpheme labels that were found in ODIN's 1,081 lines of Turkish IGT.   However, we normalized those morpheme labels throughout the entire ODIN database.  We show a few normalized examples in Table~\ref{table:normalization}.  

The Arapaho dataset was originally created in ToolBox~\cite{Toolbox}.   Cleaning of this dataset consisted of running an in-house script that found lingering formatting errors.  Some of the errors were corrected by Andy Cowell, and others remain in a default format created by our script and will be corrected in the future. 

\subsection{Data Preparation: Turkish-English NMT}
The 1,081 Turkish-English glosses are split into training, validation, and test sets with the ratio of 0.8,0.1,0.1. Our training data only contains 865 lines. We choose Turkish because it is morphologically rich \cite{matthews2018using,botha2014compositional}, agglutinative, and has words that cannot be translated as a single word in other languages 
\cite{clifton2011combining,el2006initial,bisazza2009morphological}. 

\section{Models}
\label{sec:methodology}
For convenience, we use \textit{1}, \textit{2}, \textit{3}, \textit{4} to denote each line of the
translation sequence as shown in Table~\ref{table:notation}.

\subsection{An Extension to Multilingual NMT:  Gloss-to-Target Translation}
In Figure~\ref{figure:multi}, we show a simple setup for multilingual translation on the top. Each source sentence is tagged with the source and language tags and is added to the training data. In Figure~\ref{figure:inl}, we also show our model of gloss-to-target translation, which introduces a new extension to multilingual NMT. 

In our gloss-to-target translation, we train a multilingual NMT system on 57,608 lines across 1,497 languages. Our source sentences are the interlinear glosses with target-lemma (\textit{3}), and our target sentences are the target translations (\textit{4}). We tag each glossed text with the gloss-source language tag, for example, we tag Hmong glosses with ``\texttt{blu}". As such, our training data on the source side contains ``\texttt{blu} 1SG be\_thirsty water'' (Hmong), ``\texttt{cmn} 1SG be\_thirsty'' (Chinese) and ``\texttt{deu} 1SG.ACC be\_thirsty'' (German), and our training data on the target side is ``I am thirsty'' in English.  We proceed to train using a unified attention mechanism. Even though the gloss only contains the target-lemma, our system is informed of the source language. 

In our multilingual NMT translation, we use a minibatch size of 64, a dropout rate of 0.3,
4 RNN layers of size 1000,
a word vector size of 600,
number of epochs of 13,
a learning rate of 0.8 that decays at
the rate of 0.7 if the validation score
is not improving
or it is past epoch 9. 
Our code is built on OpenNMT \cite{klein2017opennmt} 
and we evaluate our models using
BLEU scores \cite{papineni2002bleu}, and 
qualitative evaluation. 

We train a baseline attentional NMT model without adding the source language tags, as well as other system variation, some of which include Arapaho and ODIN in Section~\ref{sec:results}. 

\subsection{Case Study: Turkish-English Translation}
We use our gloss-to-target model above as the third step in our Turkish-English Translation pipeline. We present a case study of Turkish-English translation using 865 lines of training data. 

\subsubsection{\textit{1}$\rightarrow$\textit{2}: Generation of Interlinear Gloss with Source-Lemma}
We use a morphological analyzer \cite{oflazer1994two}
to generate morphological tags and a root for each word token in
the source text. After normalizing the morpheme labels, we produce an interlinear gloss with source-lemma. In Table~\ref{table:normalization2}, we show the interlinear gloss with source-lemma in every second line. 

\subsubsection{\textit{2}$\rightarrow$\textit{3}: Generation of Interlinear Gloss with Target-Lemma} 
We use a dictionary produced by aligning parallel
corpora to construct interlinear gloss
with target English tokens from source Turkish tokens \cite{dyer2013simple}.
In order to produce a higher quality dictionary, instead of choosing the 865 lines training data to construct alignments, we use an additional parallel corpus with 57,608 lines. This data, which is only used to created a dictionary, would be unnecessary if a high-quality dictionary already existed. Using the dictionary, we generate interlinear glosses with target English tokens as shown by every third line in Table~\ref{table:normalization2}.

\subsubsection{\textit{3}$\rightarrow$\textit{4}: Training NMT system for Gloss-to-Target Translation}
We use our multilingual gloss-to-target NMT model trained above to translate from glosses with target-lemma (\textit{3}) into the target language (\textit{4}). 

\section{Results}
\label{sec:results}
\subsection{Multilingual Gloss-to-Target NMT}
In the description of our results, we will use a number of labels. 
In Table~\ref{table:odin}, we use \textit{Turkish} to denote the baseline translation system (taking all the interlinear glosses with target lemma (\textit{3}) in our 865 lines of Turkish data and their target translations (\textit{4}) and training a single-source single-target translation system). We use \textit{ODIN} to denote the baseline translation system in which we take all the interlinear glosses with target lemma (\textit{3}) in ODIN and their target translations (\textit{4}) and train a single-source single-target translation system. We also use \textit{Arapaho} to denote the baseline translation system of taking all the interlinear glosses with target lemma (\textit{3}) in Arapaho and their target translations (\textit{4}) and train a single-source single-target translation system. We use \textit{ODIN+Arapaho} to denote the single-source single-target NMT model trained on both the ODIN and Arapaho datasets. 
We use \textit{ODIN\_multi} to denote the multilingual NMT model trained for gloss-to-target translation by tagging each gloss with its source language labels, for example ``\texttt{blu}'' for Hmong. We use \textit{ODIN+Arapaho\_multi} to denote the multilingual NMT model produced by tagging each gloss with its source language labels combining both the ODIN and Arapaho datasets. 

We see that \textit{ODIN\_multi} raises the BLEU score to 23.05, an increase of +2.76 compared to the baseline \textit{ODIN} as shown in Table~\ref{table:odin} and Table~\ref{table:3to4}. We notice that the translation from the interlinear gloss to target language is very good, with high accuracy in named entities translated as well as the verbs. In our second example, our model is able to predict ``store'' when there is a typo in the gold translation. In other words, our translation sometimes beats the gold translation. 

Since we have a large number of interlinear glosses in the Arapaho dataset, we also consider the case where it is added to the training set. After adding the Arapaho data to \textit{ODIN}, our \textit{ODIN+Arapaho} raises the BLEU score to 25.94, an increase of +5.65 over that of \textit{ODIN}. After adding the Arapaho data to \textit{ODIN\_multi}, our \textit{ODIN+Arapaho\_multi} has a BLEU score of 25.85, a increase of +2.80 over that of \textit{ODIN\_multi}. However, the BLEU score of \textit{ODIN+Arapaho\_multi} is lower than that of \textit{ODIN+Arapaho} by -0.09, although it is a small difference. We think a contributing factor to the similar performance of \textit{ODIN+Arapaho} and \textit{ODIN+Arapaho\_multi} is because the Arapaho dataset, though it is relatively large, is monolingual. Even though we train in a multilingual fashion in \textit{ODIN+Arapaho\_multi}, most of the training data is skewed towards the monolingual Arapaho data, therefore its performance is similar to that of \textit{ODIN+Arapaho}. 

\subsection{Case Study: Turkish-English NMT}
We use a few labels in the descriptions of our experiments. 
In Table~\ref{table:quantitative} and Table~\ref{table:qualitative},
\textit{Baseline1} denotes an attentional NMT system that trains on the 865 lines of Turkish-English parallel data without using any information from the the interlinear gloss; \textit{Baseline2} denotes attentional NMT system that trains on an additional 57,608 lines of Turkish-English parallel data; \textit{Generation} denotes the interlinear gloss with target-lemma generation step (\textit{1}$\rightarrow$\textit{2}$\rightarrow$\textit{3}). 

We use \textit{IGT\_src} to denote our translation through  gloss with source-lemma as a pivot into target language (\textit{1}$\rightarrow$\textit{2}$\rightarrow$\textit{4}). We use \textit{IGT\_tgt} to denote our translation through gloss with target-lemma as a pivot into target language (\textit{1}$\rightarrow$\textit{2}$\rightarrow$\textit{3}$\rightarrow$\textit{4}). 

For evaluation, the BLEU score does not suffice for evaluating our translation using only 865 lines of data, especially when our translating goal is to improve meaningful translation and improve fine-grained translation performance.

We present five metrics that we have designed for our evaluating purpose on top of both 4-gram and 1-gram BLEU scores. They are: noun-match accuracy, verb-match accuracy, subject-object agreement accuracy, and tense-match accuracy. The noun-match accuracy is the percentage correctly predicted string-matched nouns; and the verb-match accuracy is the percentage correctly predicted string-matched verbs. The subject-verb agreement accuracy is the percentage of correctly predicted subject-verb agreement, for example, ``he''  is matched with ``talks'' or ``talked'' rather than ``talk''. The tense-match accuracy is the percentage of correctly predicted tense. The metric of non-repetition is designed especially for NMT outputs. A lot of NMT models achieve high BLEU scores by creating spurious repetitions. To take care of this blind spot of BLEU evaluation, we introduce a non-repetition metric, the percentage of unique words in the sentence. All metrics are averaged over the entire test data and are presented as percentages in Table~\ref{table:quantitative}. These metrics were applied through hand evaluation to minimize noise; however, the process can be automated (using, e.g., POS tagging). We also evaluate using both 4-gram and 1-gram BLEU, where 1-gram BLEU helps us to evaluate our translation without considering the correct grammatical order. 

The model \textit{baseline2} performs better than \textit{baseline1} on verb-match accuracy, subject-verb agreement accuracy, and tense-match accuracy, but is worse off in BLEU scores as well as the metric of matching nouns. This is interesting because we expect that our performance will increase with increased amount of data. However, it is worth noting that linguistic gloss data is very domain specific. The Turkish ODIN dataset has a relatively narrow domain which may not be covered by the parallel data that is injected. Therefore, adding more data may not help with the translation. 

Our model \textit{IGT\_src} 
beats \textit{baseline1} and \textit{baseline2} in all metrics excluding subject-verb agreement. The reason that the subject-verb agreement does not perform well in \textit{IGT\_src} is because \textit{IGT\_src} is a factored model created by combining source text with morphological tags. Though source lemmas are tagged with subject-verb agreement information, the model finds it hard to learn about the target lemmas. \textit{IGT\_tgt} addresses this issue. The metric of non-repetition performs very well, beating all baselines. 
Our model \textit{IGT\_tgt} raises the metric of matching nouns by +20.15, raises the metric of matching verbs by +15.74, increases the metric of noun-verb agreements by +44.44, raises the metric of matching tense by +26.85, raises the 1-gram BLEU by +0.20, and raises the 4-gram BLEU by +1.66 comparing with \textit{baseline1}. It also beats \textit{baseline2} in all metrics. The reason that the noun-verb agreement performs very well in \textit{IGT\_tgt} is that the model is actively learning information of the target lemma. For example, if ``he-3.SG walks-3.SG'' is present in training, then the model learns the noun-verb agreement in the space of target lemma very well. The metric of non-repetition performs very well showing significant improvement over all baselines. 

Our five metrics only evaluate individual sentences, but there are corpus-level patterns that are also worthy of comment. For example, in \textit{baseline2}, ``the book'' is repeated across all translations even when the source sentence is totally unrelated.

In extreme low-resource scenarios like ours, qualitative evaluation is more important than quantitative evaluation. In Table~\ref{table:qualitative}, we
see clearly that the two baseline NMT systems are hallucinating. The baseline
translations have nothing in common with the source sentence, except fluency. The model \textit{IGT\_tgt} also hallucinates as it is exposed to very little information regarding the target lemmas during training. 
However, our model \textit{IGT\_tgt} produce meaningful translations that preserves the content of the source sentence to a certain extent while also achieving fluency through a good gloss-to-target NMT system. 

\section{Conclusion}
\label{sec:conclusion}
We present the cleaned and normalized Arapaho and the ODIN datasets and our multilingual model in translating from interlinear glosses to fluent target language. In addition, we present a three-step solution to extremely low-resource translation training only on 865 lines of data with linguistic information as a case study. Finally, we present five metrics for evaluating extremely low-resource translation and show that our NMT system performs well in noun-verb agreements. 

We would benefit from a more detailed gloss normalization process. We also would like to explore disambiguation in a morphological analyzer \cite{shen2016role} and more detailed morpheme segmentation.  Furthermore, IGT is ubiquitous in linguistics publications and lecture notes.  Future work could increase the size of ODIN by including IGTs from newly available publications.    

\bibliography{acl2020}
\bibliographystyle{acl_natbib}

\end{document}